%% file: acl2023.tex
\pdfoutput=1

\documentclass[11pt]{article}

\usepackage[]{ACL2023}

\usepackage{times}
\usepackage{latexsym}
\usepackage[shortlabels]{enumitem}

\usepackage[T1]{fontenc}

\usepackage[utf8]{inputenc}
\usepackage{amsmath,amsfonts,amssymb}

\usepackage{microtype}
\usepackage{multirow}
\usepackage{subcaption}
\usepackage{inconsolata}

\usepackage{graphicx}
\usepackage{amsmath}

\usepackage{makecell}
\usepackage{booktabs}
\usepackage{cleveref} %

\title{Revisiting Sentence Union Generation as a Testbed for Text Consolidation}

\author{
  Eran Hirsch$^1$ \quad
  Valentina Pyatkin$^1$ \quad
  Ruben Wolhandler$^1$ \quad \\
  \textbf{Avi Caciularu}$^1$ \quad
  \textbf{Asi Shefer}$^2$ \quad
  \textbf{Ido Dagan}$^1$ \quad \\
  $^1$ Bar-Ilan University \qquad $^2$ One AI \\
   {\small \tt \quad eran.hirsch@biu.ac.il \qquad dagan@cs.biu.ac.il} \\  
}

\begin{document}
\maketitle

\newcommand{\valentina}[1]{\textcolor{brown}{[Valentina: #1]}} 
\newcommand{\eran}[1]{\textcolor{blue}{[Eran: #1]}} 
\newcommand{\ruben}[1]{\textcolor{red}{[Ruben: #1]}} 
\newcommand{\avi}[1]{\textcolor{blue}{[Avi: #1]}}
\newcommand{\avichange}[1]{\textcolor{blue}{[Avi: #1]}}
\newcommand{\asi}[1]{\textcolor{red}{[Asi: #1]}} 
\newcommand{\ido}[1]{\textcolor{red}{[Ido: #1]}} 
\newcommand{\change}[1]{\textcolor{blue}{[Change: #1]}}

\input{content/0-abstract}
\input{content/1-introduction}

\input{content/2-relatedWork}
\input{content/3-taskDefinition}

\input{content/4-dataset}
\input{content/5-datasetQualityEvaluation}
\input{content/6-models}

\input{content/7-evaluationProtocols}

\input{content/8-results}
\input{content/9-conclusions}
\input{content/10-limitations}

\input{content/11-ethics}
\input{content/12-acknowledgements}

\bibliography{anthology,custom}
\bibliographystyle{acl_natbib}

\input{content/13-appendix}

\end{document}

%% file: content/0-abstract.tex
\begin{abstract}

Tasks involving text generation based on multiple input texts, such as multi-document summarization, long-form question answering and contemporary dialogue applications, challenge models for their ability to properly \textit{consolidate} partly-overlapping multi-text information.
However, these tasks entangle the consolidation phase with the often subjective and ill-defined content selection requirement, impeding proper assessment of models' consolidation capabilities. 
In this paper, we suggest revisiting the \textit{sentence union} generation task as an effective well-defined testbed for assessing text consolidation capabilities, decoupling the consolidation challenge from subjective content selection.
To support research on this task, we present refined annotation methodology and tools for crowdsourcing sentence union, create the largest union dataset to date and provide an analysis of its rich coverage of various consolidation aspects.
We then propose a comprehensive evaluation protocol for union generation, including both human and automatic evaluation. 
Finally, as baselines, we evaluate state-of-the-art language models on the task, along with a detailed analysis of their capacity to address multi-text consolidation challenges and their limitations.\footnote{Our data and code is available at: \url{https://github.com/eranhirs/sentence_union_generation}}

\end{abstract}

%% file: content/1-introduction.tex
\section{Introduction} 

\input{figures/figExampleUnions}

In order to acquire knowledge on a new subject or find answers to complex questions, it is often necessary to consult multiple sources of written information. While information provided in a single document is usually consistent, textual materials from various sources often use different language expressions, which may vary in terms of level of specificity, to convey similar information. An illustration of this phenomenon can be seen in Figure \ref{fig_example_unions}. In this paper, we aim to address the process of combining such multiple partially overlapping textual sources into a single unified and comprehensive format, to which we refer as \textit{text consolidation}.

Text consolidation plays a crucial role in almost any text-based information access application, such as Multi-Document Summarization (MDS) \citep{fabbri2019multinews, giorgi2022exploring}, long-form question answering \citep{fan-etal-2019-eli5, nakano2022webgpt}, and contemporary dialogue applications \citep{thoppilan2022lamda, openai2023gpt4}. It is important to point out here that content selection and consolidation manifest two distinct sub-tasks in such applications, where the former involves identifying the sought information in the source texts, based on considerations such as salience and user needs. Consolidation, on the other hand, involves merging the selected information into a coherent output text. Accordingly, we suggest that each sub-task deserves separate investigation, while focusing in this paper on the consolidation task, manifested as information union. This approach enables targeted investigation of information union capabilities of models, while enabling  modular architectures, where an effective information consolidation model can be paired with different content selection models and strategies, whether fully-automatic or interactively involving a user in the loop.

To achieve a more controlled research environment, a sentence fusion task was introduced, which fuses a set of sentences into a single sentence \citep{barzilay-etal-1999-information, thadani-mckeown-2013-fusion, agarwal2022msc}.
However, being similar to summarization, the general sentence fusion task is ill-defined, because it allows for \textit{subjective} salience-based content selection decisions \citep{daume-iii-marcu-2004-generic, krahmer-et-al-2008query-based}.
In contrast, the sentence union generation task is strictly defined as generating a sentence that contains \textit{exactly all} information from the source sentences (see Fig. \ref{fig_example_unions}).
While identifying the union task to be more attractive due to its more \textit{objective} and semantically challenging nature, we found that datasets for this topic are relatively scarce \citep{mckeown-etal-2010-time, geva-etal-2019-discofuse, lebanoff-etal-2020-understanding}, none of them sufficiently addressing the text consolidation setting.

Consequently, we revisit the sentence union generation task and propose that it can be used as an effective generic testbed for text consolidation.
Compared to the sentence intersection task, the union task is more challenging, as it requires merging both joint and disjoint information in the output and hence provides a more complete testbed for text consolidation.
Our input format is rich and challenging enough, as shown in our analyses, to support research on information merging models. Further, this setting may already be of practical use for downstream text generation tasks, for example when combined with sentence compression or decontextualization models.

Our contributions are outlined as follows: (1) we suggest focusing on sentence union generation as a resource for studying cross-text consolidation capabilities, and point out that properly identifying informational relations between pairs of sentences is necessary for proper consolidation;
(2) we provide the largest union fusion dataset to date,
while proposing a controlled annotation protocol and interface for careful creation of a sentence union corpus;
(3) we suggest evaluation protocols to assess the quality of a generated sentence union, accompanied by automatic metrics that can be used for comparing multiple systems;
(4) we provide empirical results on the abilities of prominent neural generative models to address the union task, assessing their capabilities and limitations.

%% file: figures/figExampleUnions.tex
\begin{figure}[t]
    \centering
    \includegraphics[width=1.0\columnwidth]{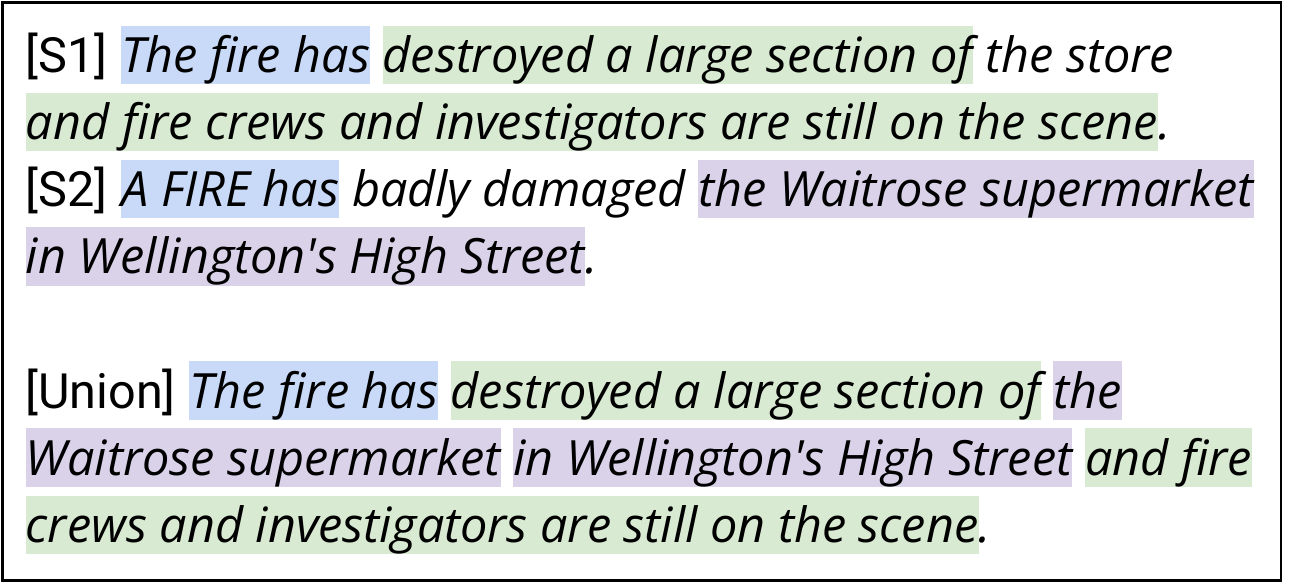}
    \caption
    {
    An example of a sentence pair and its union sentence.
    Information that must be included in the union is highlighted differently for each sentence (\textit{green} and \textit{purple} for sentences 1 and 2, respectively), unless the information is paraphrastic (equivalent) between the two sentences, which is then highlighted by the same color (\textit{blue}).
    Non-highlighted information indicates that there is corresponding information in the other sentence that is more specific.
    }
    \label{fig_example_unions}
    \vspace{-5mm}
\end{figure}

%% file: content/2-relatedWork.tex
\section{Background}

In Multi-Document Summarization (MDS) \citep{narayan-etal-2018-xsum, fabbri2019multinews} multiple-texts are summarized into a single, shorter text.
In a more controlled variant of MDS, the task requires the fusion of partly-overlapping sentences \citep{barzilay-etal-1999-information, thadani-mckeown-2013-fusion, agarwal2022msc}.
Generally, the sentence fusion task included a saliency detection (or importance) component which requires identifying which pieces of information to preserve in the fused output. As a result, sentence fusion is generally ill-defined, as different possible content selections may be valid, making the task subjective to varying necessities of a user \citep{daume-iii-marcu-2004-generic, krahmer-et-al-2008query-based}.
Its output could be seen as covering a ``loose'' intersection of the content of two sentences.

\citet{mckeown-etal-2010-time} on the other hand, to ensure more consistent fusion settings, makes a distinction between two strict variants of the task: sentence intersection and sentence union generation.
Given two (or a set of source sentences), their intersection is a sentence that contains only information that is \textit{common} to both source sentences, while their union is a sentence that contains \textit{all} information from the source sentences.
As we will see in \S\ref{sec_task_definition}, these tasks can indeed be formulated in strict entailment terms.
\citet{mckeown-etal-2010-time} crowdsourced a dataset of 300 examples for sentence intersection and sentence union, but subsequent works mostly focused on the intersection fusion part of the dataset \citep{thadani-mckeown-2011-towards, fuad-etal-2019-neufuse}.
Further, their dataset size is relatively small and primarily intended for evaluation purposes, making it inadequate for partitioning into a training dataset for fine-tuning large language models.

While \citet{mckeown-etal-2010-time} used similar sentences, whose contents partly overlap, as input, later works researched the union of disparate sentences \citep{geva-etal-2019-discofuse, lebanoff-etal-2021-modeling} where contents are disjoint. This does not address the challenge of consolidating partly overlapping texts.
In this work, we chose sentence union as a more complete testbed for multi-text consolidation.
We see our work as a continuation of the work by \citet{mckeown-etal-2010-time}, and complementary to works that introduced fusion datasets for disparate sentences.

Our work further relates to a line of research that focuses on objective generation of text.
\citet{castro-ferreira-etal-2020-webnlg} introduced a data-to-text generation task, wherein knowledge graph triplets describing facts are transformed into natural language text.
While there are many possible realizations of the knowledge graph into natural language, the task is semantically objective, with respect to the informational content expected in the output, and is hence similar to the sentence union task.
Recently, \citet{slobodkin2022controlled} introduced a new \textit{controlled text reduction} task: given an input document with highlighted spans, the task is to generate a summary in which only the information covered in the highlighted spans is included, which could be compared to a highlight union task.
Compared to our work, the spans that they used all appear in a single document, which makes it more similar to datasets which fuse disparate sentences.

%% file: content/3-taskDefinition.tex
\section{Task Formulation}
\label{sec_task_definition}
\input{figures/figRequiredLinks}

The input for our sentence union task consists of two related sentences whose content partly overlap. The output union is then defined as a single sentence that follows two conditions: (a) it contains exactly the information from the two input sentences, and (b) it does not include any redundancies in its content.
Condition (a) implies that there cannot be any information missing from the union that is mentioned in the source sentences, while at the same time the union cannot contain information that is not mentioned in the source sentences (i.e., hallucinations).
Condition (b) implies that the union must avoid repetition of any units of information stemming from the source sentences, even if they are conveyed in different lexical terms.

Notably, the semantic content of the output union (condition (a)) can be defined objectively in strict textual entailment terms.
Formally, given an input of two related sentences $s_1$ and $s_2$, and their union $u$, $u$ should satisfy $u \models s_1$ , $u \models s_2$ and $s_1 + s_2 \models u$, where $ \models $ denotes textual entailment and $ + $ denotes concatenation of the two sentences.
This definition, however, does not cover condition (b) of avoiding redundancies.

Identifying relevant informational links is crucial for producing a union, as demonstrated by the example in Fig.~\ref{fig_required_links}. We observe three types of relations between information units in the source sentences that affect the content of the resulting unit: (1) equivalent content, (2) uni-directional entailing content, and (3) disjoint content.
Equivalent content, such as lexical equivalence or paraphrases, needs to be identified and included exactly once in the union to avoid redundancy. Uni-directional entailing content pertains to aligned text spans where one span can be implied from the other. In this case, only the entailing text unit should be included: including both spans would be redundant, while including only the less specific mention would result in missing information. Disjoint content must be included in the union as it provides distinct information not mentioned in the other sentence.
For example, in Fig.\ref{fig_required_links}, sentence 1 mentions the reason for firing Weightman while sentence 2 mentions that Harvey resigned, each providing distinct information. In addition, according to our annotation scheme, we assume that the date of the publication is known, which means that when a phrase such as ``the previous Thursday" is mentioned, we can infer the specific date.
Thus, the text spans ``On March 1st'' and ``the previous Thursday'' are equivalent, while ``Francis Harvey'' in sentence 1 is more specific than the text span ``Harvey'' in sentence 2. By considering these three types of relations, a proper union can be produced.

As noted earlier, we see the union generation task as a more comprehensive setup for information consolidation than the \textit{intersection} generation task\footnote{The information content for the intersection task can also be defined in strict textual entailment terms. Formally, for the intersection $i$ of the two sentences $s_1$ and $s_2$, it is required that $s_1 \models i$ , $s_2 \models i$ and for all $i^*$ such that $s_1 \models i^*$ , $s_2 \models i^*$ , then $i \models  i^*$.}.
This is because the union output should combine all the content from both source sentences, while the output of the intersection task does not include information mentioned in only one of the sentences.
As a result, the union is more informative than the intersection, which makes it more representative for downstream multi-text tasks requiring information consolidation, aiming to create an efficient, non-repetitive output text.

%% file: figures/figRequiredLinks.tex
\begin{figure*}[t]
    \centering
    \resizebox{0.8\textwidth}{!}{
        \includegraphics{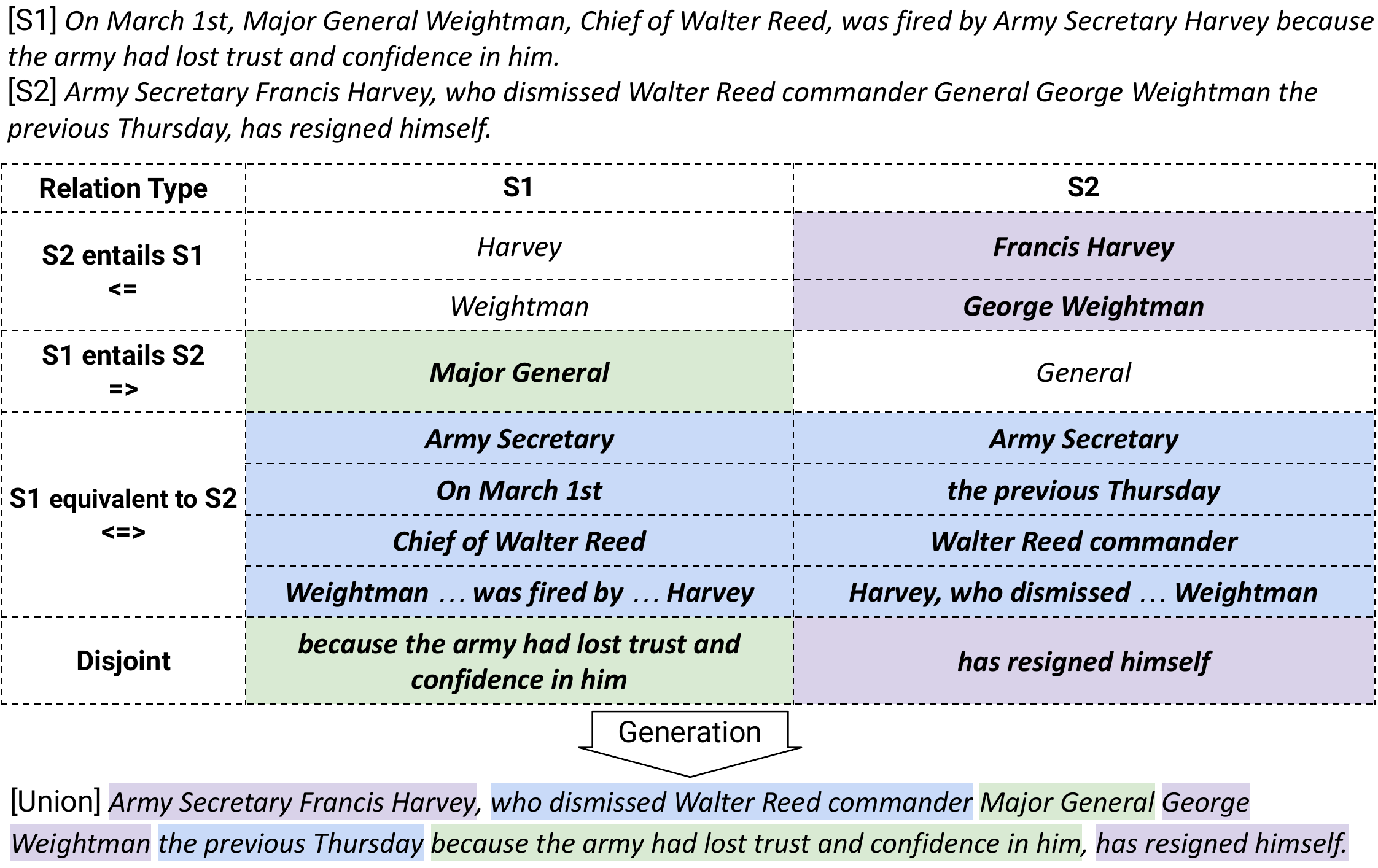}
    }
    \caption{An example of a pair of sentences, the informational relations between their text spans, and their union.
    In order to generate the union, it is first necessary to identify these relations (possibly implicitly), and then include all new or more specific information (denoted by colors) without redundancy.
    }
    \label{fig_required_links}
    \vspace{-5mm}
\end{figure*}

%% file: content/4-dataset.tex
\section{Dataset}

\subsection{Data sources}
\label{sec_data_sources}

Annotating a text consolidation sentence union dataset requires a collection of \textit{related} sentences, as input, as seen in Fig. \ref{fig_example_unions}.
Specifically, we require naturally occurring sentences with some semantic overlap, where different types of informational relations are present.
Note that we do not consider sentences with no content overlap as relevant for our dataset.

To that end, we use the dataset created by \citet{weiss-etal-2021-qaalign}, which includes pairs of relevant sentences with high semantic overlap.
Their dataset was curated by identifying information overlap between sentences, based on the repurposing of existing human annotations.
This approach is preferable to using models that identify semantic overlap, such as \citet{thadani-mckeown-2013-fusion}, since it introduces less bias to the dataset.
The original datasets from which they sourced the sentences include: (1) the Event Coreference Bank (ECB+, an extension over ECB) \citep{cybulska-vossen-2014-ecb}, which provides annotations for coreferring event and entity mentions, (2) MultiNews (MN) \citep{fabbri2019multinews}, which contains clusters of news articles along with human-written summaries, and (3) The Document Understanding Conference (DUC) and the Text Analysis Conference (TAC)\footnote{\url{https://duc.nist.gov/} , \url{https://tac.nist.gov/}}, both providing MDS evaluation datasets.

\subsection{Annotating sentence union}
\label{sec_annotating}
\input{figures/figUnionAnnotationInterface}

The process of writing a sentence union involves carefully tracking information units and blending them together to form the output, as outlined in \S\ref{sec_task_definition}.
We introduce an elaborate crowdsourcing approach and interface (see Figure \ref{fig_union_annotation_interface}) for annotating union datasets at a large scale, which splits the annotation process into multiple steps.

Starting with the two source sentences, the first step is to choose one sentence as the \textit{base sentence}, that will be used as the basis for generating the sentence union, depicted in (Fig.~\ref{fig_union_annotation_interface}, [1]).
Our early experiments have shown that it is easier to merge the information from one sentence by adding it to the other sentence than write a merged sentence from scratch.
We instruct the workers to choose the more detailed sentence as the base sentence, since this sentence would usually require less edits when merging into it information from the other sentence.
In the other sentence, termed the \textit{integrated sentence}, the worker has to highlight which spans they would like to integrate into the base sentence (Fig.~\ref{fig_union_annotation_interface}, [2]).
Finally, in the writing step, the worker blends the highlighted spans into the base sentence, thus creating the sentence union (Fig.~\ref{fig_union_annotation_interface}, [3]).

To optimize the diversity of inputs within our dataset while considering our annotation budget, each example was assigned to a single annotator.
To ensure the quality in annotators’ decisions, our process follows the controlled crowdsourcing approach \citep{roit-etal-2020-controlled}.
See App.~\ref{appendix_annotation_interface} for more details and screenshots of the entire annotation process.

\paragraph{Skipping examples}
In certain cases, it may not be possible to generate a coherent sentence union from a pair of sentences, and annotators were given the option to skip such examples. A comprehensive analysis of these skipped cases is presented in Appendix~\ref{appendix_skip}. Mainly, our findings indicate that the dataset from which we derived our data\cite{weiss-etal-2021-qaalign}, and was primarily designed for proposition alignment, contains many sentence pairs that are not sufficiently related to each other and hence are not suitable for producing a meaningful union.

\paragraph{Subtle annotation cases}
In addition to the aforementioned instructions, we took into consideration a few prominent special cases concerning the source sentences that would affect the resulting sentence union. Such cases include the need for world knowledge, temporal issues, subjectivity and attribution. For examples and guidelines provided to the workers for such cases, refer to App. \ref{appendix_subtle_annotation_cases}.
 
\subsection{Cleaning annotations}

In order to ensure a high quality dataset, we introduced a post-processing step in which we either removed or manually edited examples matching specific filtering criteria.
Filtering included finding non-overlapping input sentences based on their output union (i.e., the output was a simple concatenation of the two source sentences), as well as automatically identifying and manually reviewing subtle annotation cases described in App. \ref{appendix_subtle_annotation_cases}.
For more details, see App. \ref{appendix_cleaning_annotations}.

%% file: figures/figUnionAnnotationInterface.tex
\begin{figure}[t!]
    \centering
    \resizebox{\columnwidth}{!}{
        \includegraphics{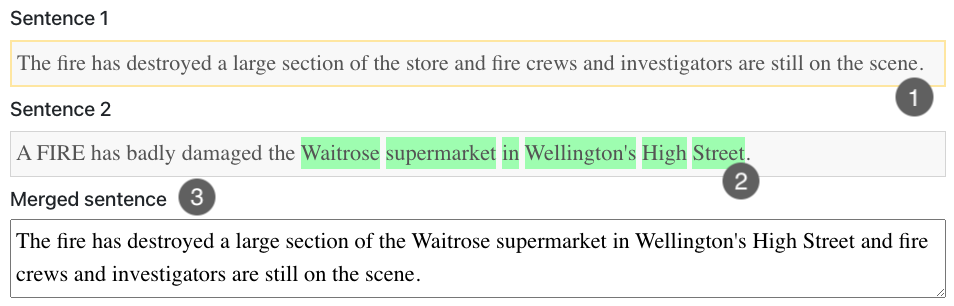}
    }
    \caption{
    A screenshot of the sentence union text generation annotation interface.
    The screenshot shows the last step, where the worker already choose sentence 1 as the base sentence [1], highlighted the new or more specific information in sentence 2 [2] and wrote the final sentence union (``Merged sentence'') [3].
    }
    \label{fig_union_annotation_interface}
    \vspace{-5mm}
\end{figure}

%% file: content/5-datasetQualityEvaluation.tex
\section{Dataset Analysis and Assessment}
\label{sec_dataset_quality_evaluation}

\input{tables/tabSizes.tex}

In the following subsections, we report various analyses of the quality and other properties of our dataset.
Dataset split statistics appear in Table \ref{tab_dataset_sizes}.
Our approach yielded a test dataset comprising of 477 instances, a sample size which is reasonable in light of the confidence intervals outlined in \S\ref{sec_results}. Moreover, our analysis of learning curves (see Appendix \ref{appendix_learning_curve}) suggests that the size of our training dataset is sufficient, and further expansion may not yield significant benefits.

\subsection{Sentence union quality}

To estimate the reliability of our dataset, we have conducted a human assessment on a sample of 100 examples of sentence unions generated by our annotators. 
Our goal is to check whether the sentences in the dataset objectively fulfill the union requirements defined in Sec.~\ref{sec_task_definition}.
For this purpose we designed two evaluation criteria for content (\textit{coverage}, \textit{faithfulness}), and one criterion for finding redundancies (\textit{redundancy}).
In addition, we evaluate the fluency of the generated sentence, as commonly done for generation tasks.

\begin{itemize}[wide, labelwidth=!, labelindent=0pt]
    \item \textbf{Coverage:} Does the sentence union contain \textit{all} information expressed in the source sentences?
    
    \item\textbf{Faithfulness:} Does the sentence union describe \textit{only} information expressed in the source sentences?

    \item\textbf{Redundancy:} Does the sentence union redundantly repeat some information?

    \item\textbf{Fluency:} Does the sentence union progresses fluently, form a coherent whole and is easy to understand?
\end{itemize}

The content criteria resemble closely those used for data-to-text generation tasks \citep{castro-ferreira-etal-2020-webnlg} which also require exact content matching between their input and output.
We add another criterion for evaluating redundancies, as our input does include redundancies which needs to be avoided in the output.

\input{tables/tabUnionQuality}

As a simple way to measure the content criteria, we count the number of content words\footnote{We removed stop words using \url{www.nltk.org}.} involved in pieces of information that are missing from the sentence union, or are unfaithful to the source sentences.
For example, if the sentence union in Fig~\ref{fig_required_links} would not mention the name \textit{``Nick Jones''}, which was mentioned in sentence 2, we count this as 2 misses.
A more complicated example would be if the sentence union attributes \textit{``Nick Jones''} to the wrong entity, such as \textit{``FBI Deputy Director Nick Jones''}.
In such case, we consider the entire span (5 words) as missing, as well as unfaithful.
Note that faithfulness can be seen as symmetrical to coverage, where we simply count content words in the sentence union that are not supported in the source sentences.
Similarly, for the redundancy score, we count the number of content words involved in pieces of information that are redundant in the union.
For example, in the phrase \textit{``Thursday overnight at 2:09am''}, the phrase \textit{``overnight''} is considered redundant, and we will count 1 redundant word.
We did not notice any fluency issues in the sentence unions created by the workers, as may be naturally expected given the high quality of our selected workers.

We start by counting the number of content words in all of the sentence unions in our sample, which adds up to 2372 content words, termed $w_{total}$.
Then, to create a \textit{coverage} score, the count of missing content words is termed $w_{missing}$, and the coverage score is calculated as $\frac{w_{total}}{w_{total}+{w_{missing}}}$.
To create a \textit{faithfulness} and \textit{redundancy}
scores, we calculate $1-\frac{w_{unfaithful}}{w_{total}}$ and $1-\frac{w_{redundant}}{w_{total}}$, respectively, where $w_{unfaithful}$ is the number of unfaithful words and $w_{redundant}$ is the number of redundant words.
Results for these metrics are available in Table \ref{tab_union_quality}.
Overall, coverage issues were encountered in 8 examples out of 100, faithfulness and redundancy issues in one example each.

\paragraph{Quality comparison to the prior dataset} We compare our dataset to the \citet{mckeown-etal-2010-time} dataset of 300 sentence unions examples.
In their annotation process, 5 workers annotated each pair of sentences, and then a single sentence union out of the 5 was automatically chosen as a representative.
We evaluated a sample of 20 such representative sentence unions and used the same quality metrics that were used in our dataset quality analysis, reported in Table \ref{tab_union_quality}.
We conclude that our controlled process, which separates the identification of informational relations from the writing phase, results in higher quality sentence unions, making significantly less coverage and redundancy mistakes, which are often due to lack of attention to details.
For the faithfulness criterion, both approaches achieved similar high scores, which is expected since humans are not prone to hallucinate when editing a sentence.
Overall, our annotation process achieves slightly better results, while employing only one worker instead of five.

\subsection{Dataset compression rate}
\label{sec_compression_rate}

\input{figures/figCompressionRate.tex}

Our motivation for the union task is to develop models that can consolidate information from naturally occurring texts with varying degrees of overlapping information.
Hence, in order to assess the diversity of our dataset with respect to the degree of such information overlap, we suggest to compute and analyze the \textit{Compression Rate} (CR) in our instances, which measures in our setting the amount of redundancies (unlike the data-to-text setting) between the two source sentences\footnote{In the union task, compression refers only to the merging of redundancies across the source sentences.}.
By design, a CR of 100\% would imply that a single source sentence contains all of the information in both source sentences, which means that the other sentence is completely redundant.
A CR of 0\% would imply that there is no redundancies between the source sentences.

Denoting our two input sentences \texttt{short} and \texttt{long}, per their lengths, as well as the \texttt{union} sentence, and following the rationale above, the compression rate is calculated as the amount of information that is eliminated from the shorter sentence.
Formally, we have $\text{CR}\left( \text{\texttt{short}}, \text{\texttt{long}}, \text{\texttt{union}}\right) = 1 - \frac{|\text{\texttt{union}}| - |\text{\texttt{long}}|}{|\text{\texttt{short}}|}$ , counting sentence length by content words.

As can be seen in Fig.~\ref{fig_compression_rate}, our dataset supplies a variety of examples in terms of CR for every split.
We report an average CR score of $60.82$\textsubscript{$\pm0.67$} for our dataset and an average CR score of $65.62$\textsubscript{$\pm1.35$} for \citet{mckeown-etal-2010-time}.
These results imply that our dataset on average contains somewhat less overlap between the source sentences, overall includes a large variety of redundancy levels.

\subsection{Informational relations analysis}
\label{sec_dataset_alignment_properties}

Complementary to the analysis in \S\ref{sec_compression_rate}, naturally occurring texts can include a wide variety of cross-text informational relations, as described in \S\ref{sec_task_definition}.
For this reason, we analyzed the frequency of the more challenging relations necessary to generate proper sentence union.
Our analysis includes a sample of 30 sentence pairs from our dataset.
On average, a sample of 10 examples is expected to include 17 ``paraphrastic uni-directional entailment'' relations (a uni-directional entailment which differs lexically), such as \textit{``supermarket''} entailing \textit{``store''}, or \textit{``gave interviews on NBC's today''} entailing \textit{``appearance on NBC's today''}.
As described in \S\ref{sec_task_definition}, such examples challenge a consolidation model to include only the \textit{entailing} expression in the output.
In addition, such a sample is expected to include 21 paraphrastic equivalence relations.
These challenge the model to include only one of the equivalent expressions in the output, to avoid repetition.
Overall, these statistics assess the abundant semantic challenges posed by our dataset.

%% file: tables/tabSizes.tex
\begin{table}[t!]
\centering
\resizebox{0.6\columnwidth}{!}{%
\begin{tabular}{l|lll|l}
\toprule
Split & Train & Dev & Test & Skipped \\
\hline
\midrule
Size &  1087 &  349 &  477 &     458 \\
\bottomrule
\end{tabular}
}
\caption{Sizes of the splits of our dataset, as well as of the skipped examples (19.3\% of \citet{weiss-etal-2021-qaalign}).}
\label{tab_dataset_sizes}
\end{table}

%% file: tables/tabUnionQuality.tex
\begin{table}[t!]
\centering
\resizebox{\columnwidth}{!}{%
\begin{tabular}{l|lll}
\toprule
Datasets & Coverage & Faithfulness & Redundancy \\
\hline
\midrule
Ours                           &   98.3\% &       99.8\% &        99.8\% \\
\citet{mckeown-etal-2010-time} &   96.5\% &       99.5\% &        98.6\% \\
\bottomrule
\end{tabular}
}
\caption{
Evaluation of union quality.
}
\label{tab_union_quality}
\end{table}

%% file: figures/figCompressionRate.tex
\begin{figure}[t]
    \centering
    \includegraphics[scale=.5]{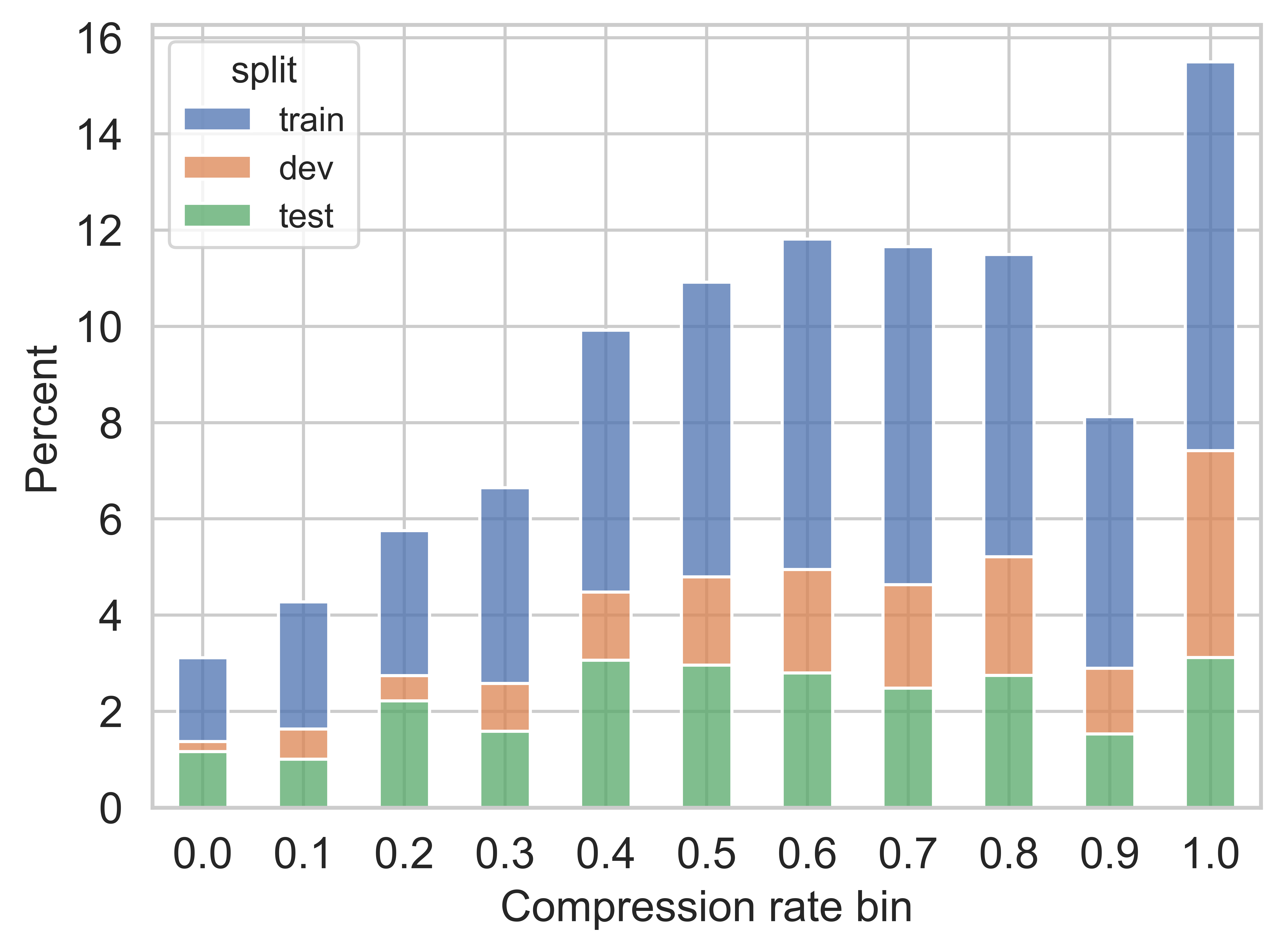}
    \caption{Compression Rate (CR) vs. the frequency of each CR bin, for the train/dev/test dataet splits.}
    \label{fig_compression_rate}
    \vspace{-3mm}
\end{figure}

%% file: content/6-models.tex
\section{Baseline Models}
\label{sec_models}
We present baseline models, aiming to test neural pretrained language models' for their ability to implicitly recognize relevant informational relations between input sentences and properly create their union.

\paragraph{Fine-tuned models}
As our first type of baseline we fine-tune a large pre-trained sequence-to-sequence model using our data.
To that end, we picked two strong models: \(T5_{large}\) \citep{raffel2019t5}, which is commonly applied to end-to-end text generation tasks~\cite{chen2020big}, and \textsc{Primera} \citep{xiao-etal-2022-primera}, which was pre-trained in a cross-document fashion \cite{caciularu-etal-2021-cdlm-cross} and achieves state-of-the-art results over multi-document summarization datasets. This makes this model appealing for our sentence fusion task, where the two sentences originate in different documents.
See App. \ref{appendix_training_details} for information about training details.

\paragraph{In-context learning}
Another current baseline approach is in-context learning, in which the instructions and examples to the task are provided as input (the prompt) at inference time to very large pre-trained language models.
We used $GPT3$ \cite{brown2020language}, specifically \textit{text-davinci-003}.
The instructions we initially used were similar to those given to the annotators. %
We then optimized the prompt by running it on the training dataset and manually identifying mistakes. The identified mistakes were added to the prompt as examples.
In addition, we added to the instructions ``important'' notes to what the model should pay attention to.
See App. \ref{appendix_prompting} for the complete final prompt and configuration used.

%% file: content/7-evaluationProtocols.tex
\section{Model Evaluation Protocols}
\label{sec_evaluation_protocols}

We evaluate our baseline systems both through human evaluation (\S\ref{sec_human_evaluation}) and with automatic metrics (\S\ref{sec_automatic_metrics}) suitable for the task, which can generally be used in the development cycles of union generation systems (\S\ref{sec_automatic_metrics}).

\subsection{Human evaluation}
\label{sec_human_evaluation}

\input{tables/tabEvalCriteria.tex}

The human evaluation is conducted over the predicted unions for the test set for each of the baseline models.
Instead of judging the generated sentence union for each baseline system separately, the evaluation is done in a comparative fashion, following previous works where the evaluator sees together the outputs of all baseline systems \citep{callison-burch-etal-2007-meta, novikova_2018_rankme}.

Similar to the analysis of the dataset quality in \S\ref{sec_dataset_quality_evaluation}, we are interested in evaluating the coverage, faithfulness, redundancy and fluency of the predicted union, this time in a manner that fits crowdsourced human evaluation.
Content and redundancy are scored on a scale from 1 to 4 (higher is better), described in Table \ref{tab_eval_criteria}.
This scale is inspired by the Semantic Textual Similarity human evaluation approach \citep{agirre-etal-2013-sem}, which also tests for information overlap.
For the fluency score, we use a common Likert scale from 1 to 5 \citep{fabbri-etal-2021-summeval}.
See App. \ref{appendix_evaluation_process} for details and screenshots.

As there exist trade-offs between the two content measures and the redundancy measure, we add an additional measure which evaluates \textit{consolidation} as a whole.
For example, by arbitrarily adding more information to the union we can increase the coverage, but also risk increasing redundancies and unfaithfulness.
The \textit{consolidation} measure simply averages the three aforementioned measures, thus testing for overall text consolidation quality.

\subsection{Automatic evaluation}
\label{sec_automatic_metrics}

In line with previous works in text generation, we report the ROUGE metric between the reference union and the predicted union.
However, like for most generation tasks, ROUGE will unfairly penalize correct but paraphrastic sentence unions (as described in \S\ref{sec_task_definition}).
To partly address this issue, we add another automated metric which tests for bi-directional textual entailment (aka NLI), comparing the reference union sentence to the predicted union sentence, requiring entailment in both directions.
Specifically, we use the $DeBERTa_{xxlarge} v2$ model \citep{pengcheng2020deberta}, fine-tuned with the MNLI task \citep{williams2017mnli} and a threshold of 0.5.

While both metrics test for content matching, they would not penalize a model that bluntly concatenates the two input sentences. %
Therefore, we also report $\Delta{CR}$ (\S\ref{sec_compression_rate}), calculated as the average  difference between the CRs of the predicted vs. the reference union sentences (the latter is subtracted from the former), on each instance. 
A positive value thus indicates that the model compression rate is higher than that of the reference union, while a negative value indicates the opposite (model compresses less than the reference).

%% file: tables/tabEvalCriteria.tex
\begin{table}[t!]
\centering
\resizebox{1.0\columnwidth}{!}{%
\begin{tabular}{l|ll}
\toprule
Score & Content & Redundancy \\
\hline
\midrule
1 & Substantial information is missing. & Substantial information is repeated. \\
2 & Some information is missing. & Some information is repeated. \\
3 & Minor details are missing. & Minor details are repeated. \\
4 & Nothing is missing. & Nothing is repeated. \\
\bottomrule
\end{tabular}
}
\caption{The ordinal scales used for the content (coverage \& faithfulness) and redundancy measures.}
\label{tab_eval_criteria}
\vspace{-3mm}
\end{table}

%% file: content/8-results.tex
\section{Results and Analysis}
\label{sec_results}

\input{tables/tabResults}
\input{figures/figResultsHistograms.tex}

\subsection{Human evaluation of the models}
\label{sec_human_eval_results}

Results are presented in Table \ref{tab_results}, and example generations with their respective scores are provided in App. \ref{appendix_example_unions}.
The trade-off mentioned in \S\ref{sec_human_evaluation} between increasing coverage while still remaining faithful and without redundancies is evident in the results of $T5_{large}$ and $GPT3$.
\textsc{Primera} comes out as a slightly better model, as it achieves the highest consolidation score, with yet a lot of room for improvement.

To get a better sense of the absolute performance of the union sentences generated by the baseline models, we compare them to two naive models which output: (1) the concatenation of the source sentences (no avoidance of \textit{redundancy}), and (2) the longer sentence (no attempt to consolidate and \textit{cover} information from the other sentence).
Based on evaluation of 50 examples completed by the authors, we report an average redundancy score of 1.6\textsubscript{$\pm$.1} for the concatenation and an average coverage score of 2.3\textsubscript{$\pm$.1} for the longer sentence.
As reported below, all our baseline models outperform these naive models by a large margin.

Further, we draw a plot (Fig.~\ref{fig_results_hist}) of the minimal system score amongst the three component measures that the consolidation measure combines.
We note that even for the best model, \textsc{Primera}, only 29.7\% of the predictions are fully correct with respect to content and redundancy, another 40.6\% examples include minor errors, and 26\% examples contain substantial errors in at least one of the measures, indicating the limitations of current models.

\subsection{Automatic evaluation of the models}
\label{sec_automatic_eval_results}

While automatic metrics are clearly less reliable than human metrics, they can be useful for development cycles.
The automatic metric results are also reported in Table \ref{tab_results}, observing that both the $ROUGE1$ score
is highest for \textsc{Primera}, while the NLI score is highest for $GPT3$.
The $\Delta{CR}$ scores roughly correlate with the combination of coverage and redundancy detected in the human evaluation, where both lower coverage (undesired) and lower redundancy (desired) increase compression rate.

To identify the potential utility of our automatic metrics, we follow the standard practice \citep{fabbri-etal-2021-summeval} and calculate a Kendall $\tau$ coefficient \citep{mcleod2005kendall} between the human and automatic evaluation results.
Our results show that $ROUGE 1$ is the highest correlated metric with the consolidation measure ($\tau = 0.38$, $p < 0.05$).
Overall, these automatic metrics can be used in tandem to provide certain feedback during model development cycles.

\subsection{Error analysis}
To shed light on the various errors made by the baseline models, we examined 20 erroneous examples identified in the human evaluation, with each example consisting of three predictions, one from each of the baseline systems. Our findings indicate that the most frequent causes of model errors are related to the complexity of informational relationships present in the source sentences, with uni-directional entailment being the most common. Moreover, the models seem to face difficulties in accurately combining related information, which often results in incorrect merging of information with the wrong entity or predicate. Further details on the analysis can be found in Appendix~\ref{sec_error_analysis}.

%% file: tables/tabResults.tex
\begin{table*}[t!]
\centering
\resizebox{\textwidth}{!}{%
\begin{tabular}{l|lllll|llll}
\toprule
{} &                  \makecell{Coverage\\(1 to 4)} &              \makecell{Faithfulness\\(1 to 4)} &                \makecell{Redundancy\\(1 to 4)} &             \makecell{Consolidation\\(1 to 4)} &                   \makecell{Fluency\\(1 to 5)} &                   \makecell{$ROUGE 1$} &                         \makecell{NLI} &                 \makecell{$\Delta{CR}$} \\
\hline
\midrule
\textsc{Primera} &           \makecell{3.2} &           \makecell{3.7} &  \makecell{\textbf{3.8}} &  \makecell{\textbf{3.6}} &           \makecell{4.1} &  \textbf{89.92\textsubscript{$\pm$.4}} &          86.37\textsubscript{$\pm$1.6} &   9.28 
 \textsubscript{$\pm$1.5} \\
$GPT3$           &  \makecell{\textbf{3.5}} &           \makecell{3.5} &           \makecell{3.5} &           \makecell{3.5} &           \makecell{3.8} &           85.35\textsubscript{$\pm$.4} &  \textbf{96.23\textsubscript{$\pm$.9}} &  -8.83\textsubscript{$\pm$1.6} \\
$T5_{large}$     &          \makecell{2.8} &  \makecell{\textbf{3.8}} &  \makecell{\textbf{3.8}} &           \makecell{3.5} &  \makecell{\textbf{4.2}} &           85.88\textsubscript{$\pm$.5} &          73.38\textsubscript{$\pm$2.0} &            27.2  \textsubscript{$\pm$1.7} \\
\bottomrule
\end{tabular}
}
\caption{Human (left) and automatic (right) evaluation results of system generated unions over the complete test set. All scores are averages, along with their standard error (standard error for manual evaluation results was always smaller than 0.01, and is therefore omitted from the table).}
\label{tab_results}

\end{table*}

%% file: figures/figResultsHistograms.tex
\begin{figure}[t]
    \centering
    \includegraphics[scale=.5]{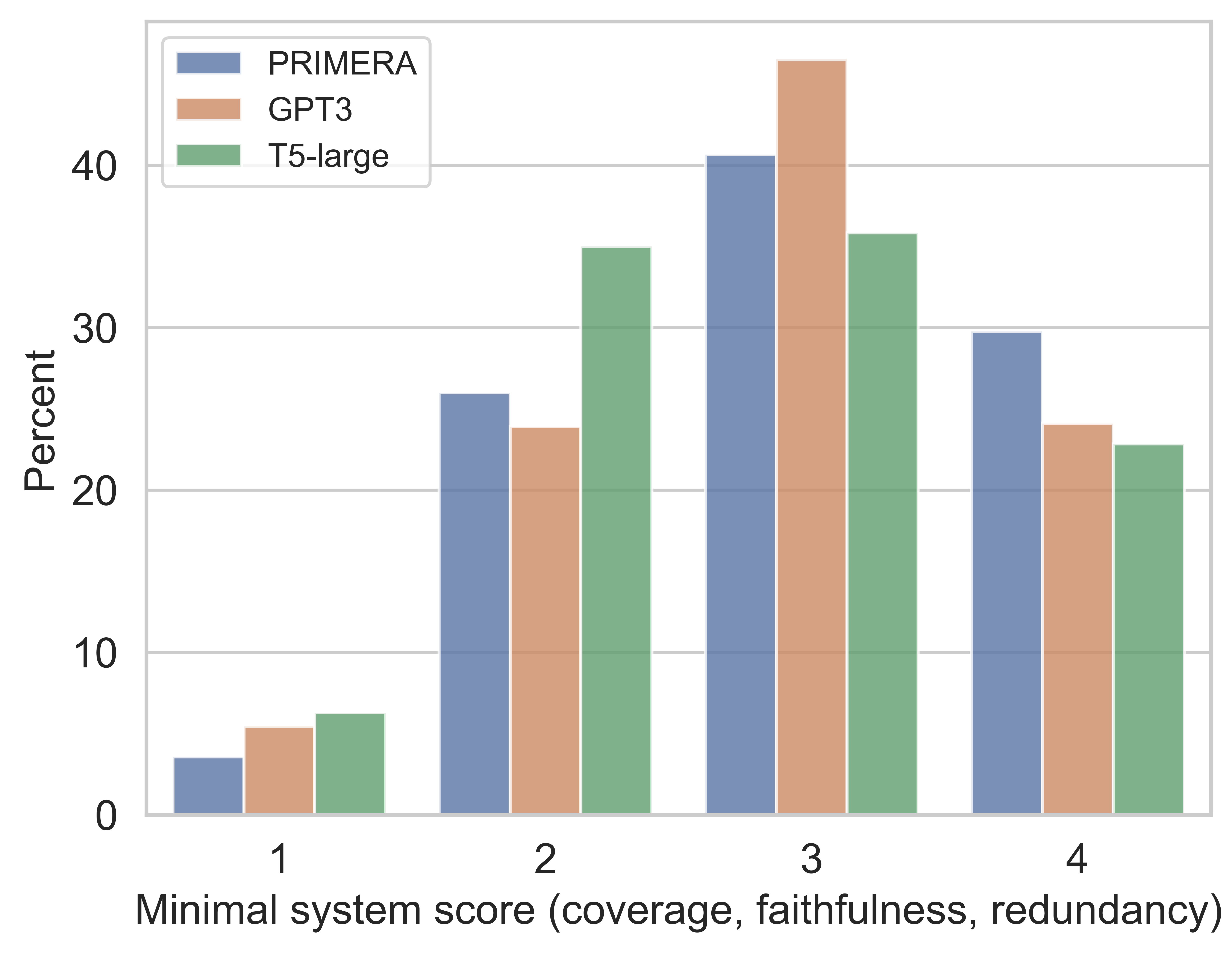}
    
    \caption{
        A histogram of minimal system scores, testing for coverage, faithfulness or redundancy mistakes.
    }
    \label{fig_results_hist}
    \vspace{-5mm}
\end{figure}

%% file: content/9-conclusions.tex
\section{Conclusions}

In this paper, we advocate for using the sentence union task as a testbed for multi-text consolidation.
We release a realistic dataset, together with a set of analyses that show that the dataset is of high quality, and challenging for multi-document consolidation efforts.
We evaluate the performance of state-of-the-art pretrained large language models on text consolidation, where our findings suggest key challenges for future research.

Future research may expand upon our dataset to include consolidation beyond 2 input sentences, and may examine the use of explicit text consolidation structures for improving multi-text consolidation in large language models.

%% file: content/10-limitations.tex
\section*{Limitations}
We enumerate some limitations to our work. While we did create the largest union dataset to date, it is still of moderate size. As shown by our learning curves (App.~\ref{appendix_learning_curve}), the amount of training data we created seemed sufficient to saturate the learning of the models with which we experimented, but it might still be found insufficient for training other models.

Our annotation protocol might have influenced the compression rates of the unions, as we instructed workers to annotate sentence unions by first choosing a base sentence and then highlighting the other sentence. Additionally, while the highlighting facilitates the annotation process, it cannot directly be used for analyses of the dataset since it is uni-directional.

The dataset includes only input with exactly two sentences and it might be desirable for future works to also be able to train systems that take more than two sentences as input. Our dataset is also domain specific, in that all the sentences are taken from news sources. This might result in challenging cross-domain generalization. 

This dataset is limited to the English language.
While the suggested annotation protocol seemingly fits other languages, the step in which words are highlighted might prove problematic for morphologically rich languages, in which a single word includes many pieces of information.
A segmentation of the text before annotation might be required.

%% file: content/11-ethics.tex
\section*{Ethics Statement}
 
\paragraph{Crowdsourcing}
    To crowdsource the dataset, we used the Amazon Mechanical Turk\footnote{\url{https://worker.mturk.com/}} (MTurk) platform.
    To participate in the first stage of recruitment, workers were required to possess the following MTurk qualifications:
    \begin{itemize}
        \item NumberHITsApproved greater than 10000
        \item PercentAssignmentsApproved greater than 98\%
        \item WorkerLocale in US, CA, AU, GB, NZ
    \end{itemize}
    Workers were paid \$0.3 for each sentence union annotation assignment, as well as a \$1.25 bonus for every 100 assignments, and \$0.4 for each evaluation assignment, as well as a \$1 bonus for every 50 assignments.
    Overall, by an average approximation of 1.8 minutes for the first assignment, and 2.4 minutes for the second assignment, their wage is expected to start from \$10 per hour and increase as the workers are more familiar with the task and start receiving bonuses.
    
    Workers were informed that the ratings they will provide will be used to evaluate artificial intelligence models which were trained on the data they annotated.

\paragraph{Dataset} The texts that workers write that are included in our dataset are limited to the information expressed in the source sentences.
The source sentences originate from the datasets mentioned in \S\ref{sec_data_sources}, which include only texts available in public news sources and were previously made available by \citet{weiss-etal-2021-qaalign}.
Our dataset does not contain information that would make it possible to reconstruct the original documents, or any human annotations, such as the summary or coreference resolution annotation, from the original datasets.

%% file: content/12-acknowledgements.tex
\section*{Acknowledgments}
The work described herein was supported in part by grants from One AI, the Israel Science Foundation 2827/21 and the Israel Ministry of Science and Technology.   
We would like to thank the workers who have annotated this dataset and we appreciate their dedication in ensuring a high level of quality.
We express our gratitude to Dr. Kapil Thadani for assisting us in retrieving his data from an earlier research endeavor.

%% file: content/13-appendix.tex
\appendix

\section{Skip Guidelines}
\label{appendix_skip}

\input{tables/tabSkipAnalysis}

In Section \ref{sec_annotating}, it was noted that there are cases where generating a union from a pair of sentences is not suitable, and workers were given the option to skip the annotation for such examples. This section outlines the specific scenarios in which workers were directed to skip examples. Eventually, our annotators skipped 458 sentence pairs from the original dataset that we used as input, as shown in Table \ref{tab_dataset_sizes}. An analysis of a sample of 30 such cases is presented in Table \ref{tab_skip_analysis}, categorized based on the criteria below. In conclusion, we found that the dataset we used as the source of our sentence pair instances, which was originally developed  by \citet{weiss-etal-2021-qaalign} for aligning predicate-argument structures (represented as question-answer pairs), includes a significant number of instances where information consolidation in the form of sentence union is mostly irrelevant.

\paragraph{No information consolidation.} One case in which workers were directed to skip examples during annotation is when there is no partially overlapping information to consolidate from two related sentences, hence their union would simply be a concatenation of the two. This case is referred to as ``No information consolidation''. An example of this scenario is when sentence 1 mentions that \textit{``Acupuncture is the ancient Chinese medical therapy technique of inserting thin, sharpened needles into specific nerve junction points of the body,''} and sentence 2 mentions a study that found \textit{``53.8 percent of the subjects who had needles inserted in four acupuncture "zones" in the ear five times a week tested free of cocaine at the end of the eight-week study period.''} In this case, there is no need to consolidate the information from the two sentences as they provide distinct pieces of information. Sentence 1 explains what is acupuncture while sentence 2 discusses a study about it.

\paragraph{Unnatural union.}
An example of an ``Unnatural union'' scenario is when unifying two input sentences would form an awkward or unnatural sentence. For instance, if the first sentence is written in the past tense and the second one in the future tense, unifying them could lead to an unnatural sentence union. As an example, consider the following sentences: \textit{``Fannie Mae's board met Sunday night to discuss Raines' future''} and \textit{``The directors of Fannie Mae, the big mortgage finance company, will meet Sunday to consider the fate of two senior executives who signed off on financial statements that violated accounting rules, people close to the company said Friday.''} Here, the first sentence uses the past tense while the second sentence uses the future tense. It would be more natural to use the past tense in the sentence union since the event occurred in the past. However, incorporating the information that someone said something on Friday before the event could result in an awkward sentence union.

\paragraph{Missing context.}
This case happens when two sentences need to be interpreted in the broader text context, which is missing in our annotation scenario, for example when there is a dangling reference to an entity that is not specified in the given sentence. This is often not problematic, unless understanding the identity of the entity is necessary to create the union. For instance, one sentence quotes a person, while the other sentence does not mention the speaker. An example of this scenario is the following: \textit{``Sadly, because Magic Leap seldom hires and does not actively recruit female candidates, the company loses competitive advantage to products like Microsoft's Hololens.''} and \textit{``When Tannen Campbell was hired by Magic Leap in 2015, the Florida company had no women in leadership roles and its only idea to make its product female-friendly was to release a pink version, according to Forbes.''} Merging these two sentences is not straightforward due to the lack of context.

\paragraph{Disagreements.} Sometimes, there are two  statements that contradict or disagree with one another.
For example, sentence 1 is \textit{``Video of Brooklyn Mother of 13 Zurana Horton shot and killed in a gang shooting was revealed Thursday .''} and sentence 2 is \textit{``A shocking video released for the first time Thursday captures the moment a Brooklyn mother of 12 was killed in a gang shootout as she picked her daughter up from school .''}.
Sentence 1 mentions that the child is 13 years old while sentence 2 mentions that the child is 12 years old.

\section{Subtle annotation Cases}
\label{appendix_subtle_annotation_cases}
\input{tables/tabSubtleAnnotationCasesAnalysis}

In Section \ref{sec_annotating} we noted that certain special cases arose when generating a union from a pair of sentences, and were included in the instructions for annotators. This section outlines the specific instructions provided to workers, with an analysis of 50 cases (Table~\ref{tab_subtle_annotation_cases_analysis}), categorized based on various criteria as described below.

\paragraph{Attribution.} One potential issue is when the source sentences make attributions to a specific source, such as a news agency. An example of this can be seen in sentence 1 \textit{``Video of Brooklyn Mother Zurana Horton being shot and killed was revealed Thursday, according to the N.Y. Daily News.''} and sentence 2 \textit{``A shocking video released for the first time Thursday captures the moment a Brooklyn mother was killed as she picked her daughter up from school.''}, where the new information in sentence 2 is attributed to the video content, rather than to the N.Y. Daily News. Another example is when a sentence contains quotes, as changing a quote to contain more information would create an unfaithful sentence union. In such cases, the workers were allowed, whenever it seemed reasonable, to attribute combined pieces of information originating from the two sentences to a reported source, even if only parts of the combined information were explicitly attributed to this source, in one of the sentences.

\paragraph{Relative dates.} Some sentences may mention a specific time relative to when the sentence was written, such as ``yesterday'' or ``Monday'', which implies that the sentence was written in the same week of the event. Workers were instructed to assume that the date of publication is known, so there is no difference between the mention of ``yesterday'' and ``Monday'', but, for example, that ``yesterday'' is more specific than ``earlier this month''.

\paragraph{World knowledge.}  In some cases, sentences may mention the same piece of information in different levels of specificity, which requires world knowledge to identify. Workers were instructed to assume common world knowledge when creating the sentence union. An example is given for Paris, which is both a city in Texas and the capital of France.

\paragraph{Before and after an event.} 
For sentences referring to events, some may differ in their time of publication compared to the event itself. Workers were instructed to use the past tense, as the sentence union is written after the event.
For example, sentence 1 mentions an event that has already happened \textit{``After leaving Alderson at 12:30 a.m. on March 3, 2005, Martha Steward declared the 5-month experience as "life altering and life affirming."''}, while sentence 2 was written before the event \textit{``US lifestyle guru Martha Stewart is expected to leave jail on Friday after a five-month sentence for a stock scandal that reinvigorated her career rather than dooming it.''}. In this case, the sentence union should be written in the past tense, as it refers to an event that has already occurred.

\section{Annotation Process}
\label{appendix_annotation_interface}

\input{figures/figAnnotationInterfaceCompleteProcess.tex}

Screenshots of the entire annotation process are depicted in Figure \ref{fig_complete_annotation_interface_process}.
Guidelines for creating sentence unions\footnote{The complete guidelines file used for training will be published upon publication.} include writing one coherent sentence, ordering the information in a stand-alone manner (as if the sentence would have been written from scratch), meaning that the writing process should not be distracted by the original split and ordering of information in the two input sentences.
To the extent possible, the sentence union should preserve the original wording of the information, but phrasing may be \textit{minimally} adjusted to create a coherent sentence union.
Each piece of information should appear only once in the sentence union.
When there is a redundancy across the two sentences, the more specific phrasing should be chosen.

The interface helps the workers to avoid making common mistakes.
For example, in order to reduce redundancies of information in the union, if a highlighted word already exists in the base sentence, both word mentions will be marked to draw the worker's attention. Another example is warning the worker when the sentence union contains non-highlighted words from the base sentence.
Also, when integrating highlighted words into the sentence union, the worker will see yellow highlights turn into green highlights.
If the worker tries to submit the annotation with yellow highlights, the system will raise an alert.

To ensure the quality in annotators’ judgements, our process follows the controlled crowdsourcing approach \citep{roit-etal-2020-controlled}, which includes a recruitment phase, two training phases accompanied by extensive guidelines, and ongoing monitoring during the annotation of the production task. Workers were allowed to participate in primary tasks only if they had completed the entire process.
Only workers who performed well on the recruitment phase were accepted to the next training phases.
The training phases were created manually, including subtle annotation cases.
After each annotation, workers were shown gold target highlights and sentence unions\footnote{Some of the authors of the paper annotated a small set of reference gold target highlights and sentence unions.} for comparison with their own output.

\section{Cleaning Annotations}
\label{appendix_cleaning_annotations}

\paragraph{Disjoint sentences} Following the skip guidelines (see App. \ref{appendix_skip}), we automatically identified examples which their sentences are mutually exclusive and their sentence union is a concatenation of the source sentences.
We find these instances by comparing content words only, since connecting the two sentences sometimes involves non-semantic lexical changes (e.g., adding a semicolon or a comma). 
Due to the fact that there is no consolidation of information in such examples, we see them unfit for a union, as mentioned in \S\ref{sec_data_sources}, and they were not included in the dataset.
We leave the automatic categorization of sentences into whether or not they are suitable for sentence unions to future work.

\paragraph{Quotes}
Following the attribution discussion in 
App. \ref{appendix_subtle_annotation_cases}, we manually reviewed examples where the union contained a quote that was not in any of the source sentences, as well as any example that had a sentence which used a first-person perspective (e.g., ``I'', ``we'', ``mine'', ``ours'', ...).

\section{In-Context Learning}
\label{appendix_prompting}

For the in-context learning approach, we used a temperature value of 0.4 and the following prompt:
\textit{In this task, you will be presented with two sentences that overlap in information, and you are tasked to merge the information of the two into a single unifying sentence without redundancies. Important: Do not omit information. Important: Do not repeat information.}

\textit{Here is an example of a correct union and a wrong union:
Sentence 1: The February assassination of former Lebanon Prime Minister Hariri put Syria under renewed pressure from the international community to abide by U.N. Security Council Resolution 1559 and withdraw its troops from Lebanon.
Sentence 2: Foreign ministers from all European Union (EU) member states, who gathered here for a meeting, on Wednesday urged Syria to withdraw its troops completely from Lebanon.
Correct union: The February assassination of former Lebanon Prime Minister Hariri put Syria under renewed pressure from foreign ministers from all European Union (EU) member states gathered for a meeting, on Wednesday to abide by U.N. Security Council Resolution 1559 and withdraw its troops from Lebanon.
Wrong union: The international community, including the European Union (EU), has put renewed pressure on Syria to abide by U.N. Security Council Resolution 1559 and withdraw its troops from Lebanon following the February assassination of former Lebanon Prime Minister Hariri.}

\textit{The union is wrong, because it does not mention that foreign ministers gathered for a meeting on Wednesday.}

\textit{Please generate a correct union to the following sentences:}

\textit{Sentence 1:} <sentence 1 goes here>

\textit{Sentence 2:} <sentence 2 goes here>

\textit{Correct union:}

\section{Training Details}
\label{appendix_training_details}

We fine-tuned $T5_{large}$ and \textsc{Primera} models for 20 epochs on a Tesla V100-SXM2-32GB GPU.
We used a hyperparameter random search strategy.
The learning rate was tuned within the range $[1e-8, 5e-5]$, while the batch size varied between $[8, 16, 32]$. We also explored the weight decay range of $[0, 0.5]$ and warump step range of $[0, 300]$.
The best model was selected based on the $ROUGE 1$ metric.\footnote{We used the HuggingFace package~\cite{wolf-etal-2020-transformers} for both fine-tuning the models and automatically evaluating them.}
The best T5 model was obtained with a learning rate of $4.3e-6$, no weight decay, no warmup steps, batch size of $32$, after $18$ epochs.
For the best-performing \textsc{Primera} model, we used a learning rate of $3.5e-6$, weight decay of $0.5$, warmup steps of $80$, batch size of 16 and selected the best checkpoint after 9 epochs.
The training time for $T5_{large}$ and \textsc{Primera} models were approximately 1 hour and 10 minutes each.

\paragraph{Input structure} When concatenating the two source sentences to insert as input for the model, we add special separator tokens to make the model aware of the sentence boundaries.
For $T5_{large}$, we separated between the source sentences in the input using a newly created special token, while for \textsc{Primera}, we used the \textit{<doc-sep>} token, which was used in the pre-training phase to separate between input source documents.

\section{Learning Curves}
\label{appendix_learning_curve}
\input{figures/figLearningCurve}

To assess the adequacy of our dataset size, we evaluated the baseline models on different subsets of our training data ($[25\%, 50\%, 75\%, 100\%]$) and various model sizes ($T5_{base}$ and $T5_{large}$).
Based on our findings (Figure \ref{fig_learning_curve}), it appears that enhancing the model size from $T5_{base}$ to $T5_{large}$ results in performance improvement.
However, the marginal benefit of increasing training dataset size may be limited, and further gains may not be significant.

\section{Evaluation Process}
\label{appendix_evaluation_process}

\input{figures/figEvaluationInterface.tex}

As explained in Section \ref{sec_evaluation_protocols}, the evaluation process involves a comparative approach, whereby all the unions of system-generated sentences are evaluated simultaneously, as shown in Figure \ref{fig_eval_interface}. The evaluation is conducted separately for four criteria. To assess the content differences between the reference union and the system union, including coverage and faithfulness, a single sentence is designated as the base sentence, and the worker is asked to evaluate the other sentence based on the amount of missing content. The reference union serves as the base sentence for evaluating coverage, while the system union is used as the base sentence for evaluating faithfulness since any information present in the system union but absent in the reference union is deemed unfaithful. In evaluating redundancy and fluency, the evaluator is only presented with the system union without the reference union.

To assess the coverage and faithfulness criteria, the workers are required to compare the generated union with the reference union, aided by red strikethroughs on words that are not included in the generated union and green highlights on words that are not included in the reference union, as illustrated in Figures \ref{fig_evaluation_coverage} and \ref{fig_evaluation_faithfulness}. For redundancy and fluency criteria, the reference union is not needed, as demonstrated in Figures \ref{fig_evaluation_repetition} and \ref{fig_evaluation_fluency}.

\section{Example Sentence Unions}
\label{appendix_example_unions}

\input{tables/tabGeneratedExamples.tex}

See Table \ref{tab_predicted_unions} for examples of sentence unions, including the sentence unions from each predicted system.

\section{Error Analysis}
\label{sec_error_analysis}

\input{tables/tabErrorAnalysisResults.tex}
\input{tables/tabErrorAnalysisExamples.tex}

In order to perform an error analysis, we analyzed 20 examples that were rated less than perfect for all metrics based on the human evaluation (see \S\ref{sec_human_eval_results}). The findings are presented in Table \ref{tab_error_analysis}, with one representative example from each subcategory included in Table \ref{tab_error_analysis_examples}. Our key observation is that models make various coverage errors as they fail to identify the uni-directional entailment correctly in the dataset. Furthermore, models make multiple coverage and faithfulness errors by incorrectly combining information and attaching it to the wrong entity or predicate.

%% file: tables/tabSkipAnalysis.tex
\begin{table}[b]
\centering
\resizebox{0.7\columnwidth}{!}{%
\begin{tabular}{lr}
\toprule
                    Category &  Count \\
\hline
\midrule
No information consolidation &     19 \\
             Unnatural union &      7 \\
                     Mistake &      3 \\
             Missing context &      1 \\
\bottomrule
\end{tabular}
}
\caption{An analysis of 30 cases that were skipped by workers during the annotation process. Among these, some were categorized as mistakes, meaning that they should not have been skipped. }
\label{tab_skip_analysis}

\end{table}

%% file: tables/tabSubtleAnnotationCasesAnalysis.tex
\begin{table}[b]
\centering
\resizebox{0.7\columnwidth}{!}{%
\begin{tabular}{lr}
\toprule
                          Category &  Count \\
\hline
\midrule
                       Attribution &   12 \\
                    Relative dates &    4 \\
                   World knowledge &    2 \\
         Before and after an event &    0 \\
No subtle case of above categories &   34 \\
\bottomrule
\end{tabular}
}
\caption{Distribution of subtle annotation cases in a sample of 50 instances (some instances belong to more than one category).}
\label{tab_subtle_annotation_cases_analysis}

\end{table}

%% file: figures/figAnnotationInterfaceCompleteProcess.tex
\begin{figure*}[t]
    \centering
    \subcaptionbox{Step 1}{
        \includegraphics[width=0.45\textwidth]{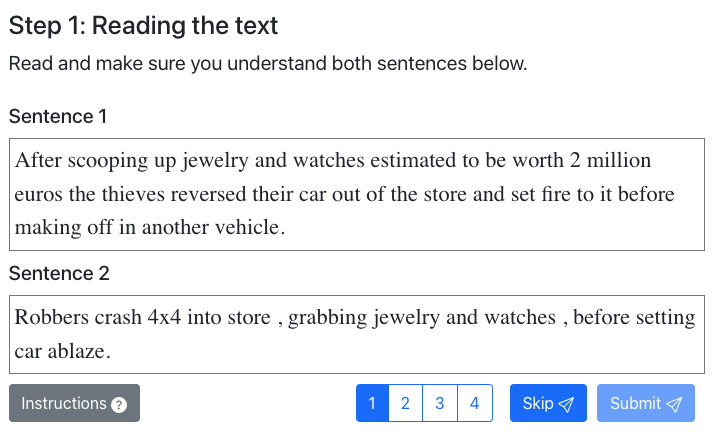}
    }%
    \subcaptionbox{Step 2}{
        \includegraphics[width=0.45\textwidth]{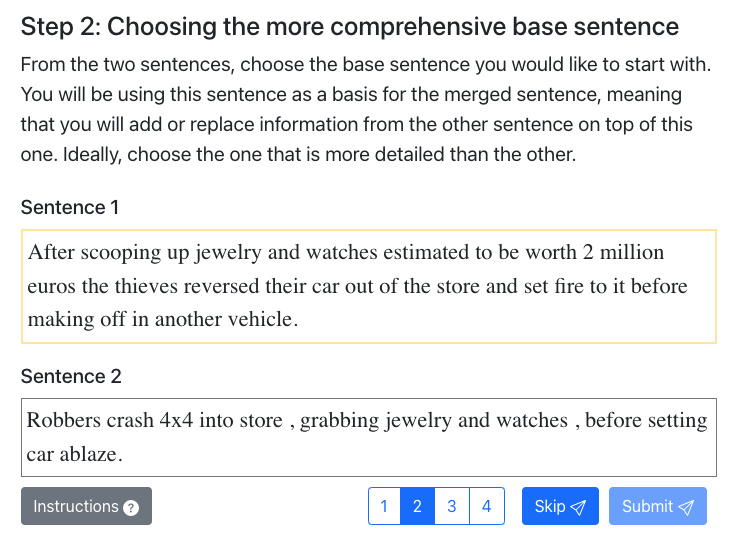}
    }\par\bigskip%
    \subcaptionbox{Step 3}{
    \centering
        \includegraphics[width=0.45\textwidth]{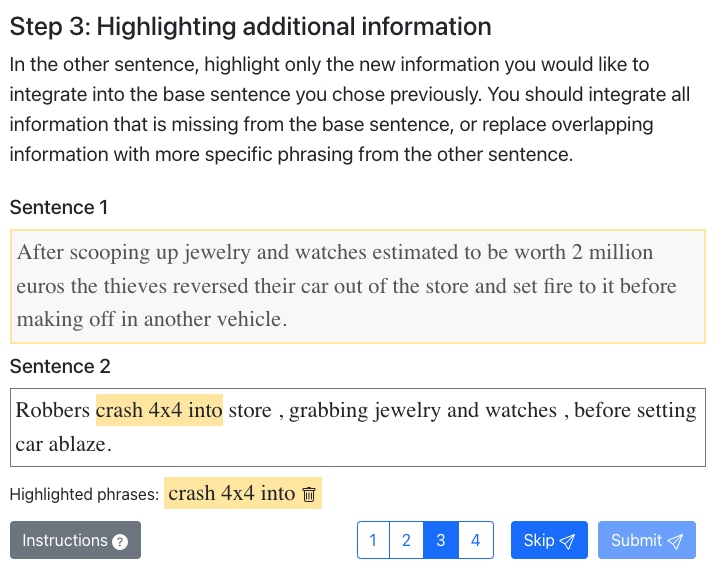}
    }%
    \subcaptionbox{Step 4}{
        \centering
        \includegraphics[width=0.45\textwidth]{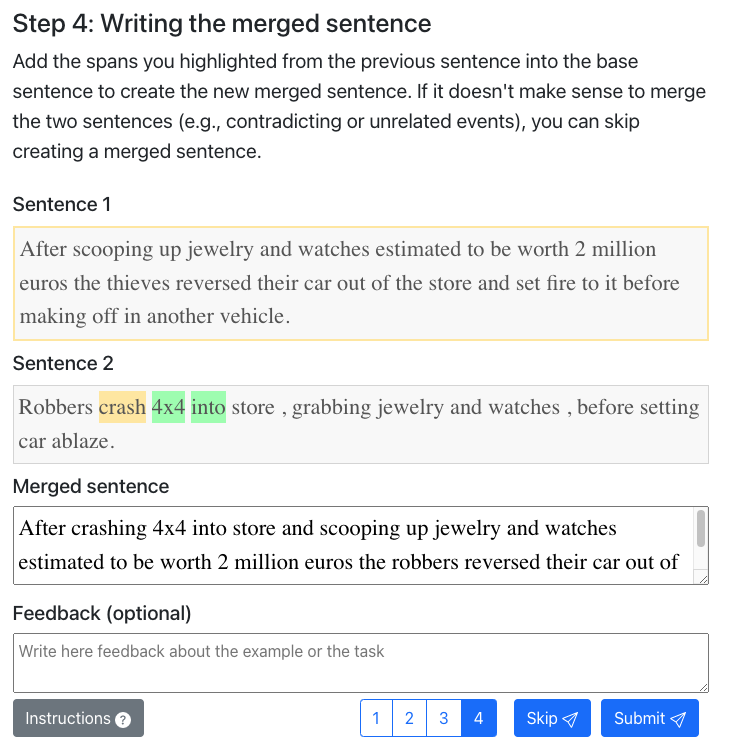}
    }%
    
    \caption{
        The interface used for the annotation process.
    }
    \label{fig_complete_annotation_interface_process}
\end{figure*}

%% file: figures/figLearningCurve.tex
\begin{figure}[t]
    \centering
    \includegraphics[scale=.5]{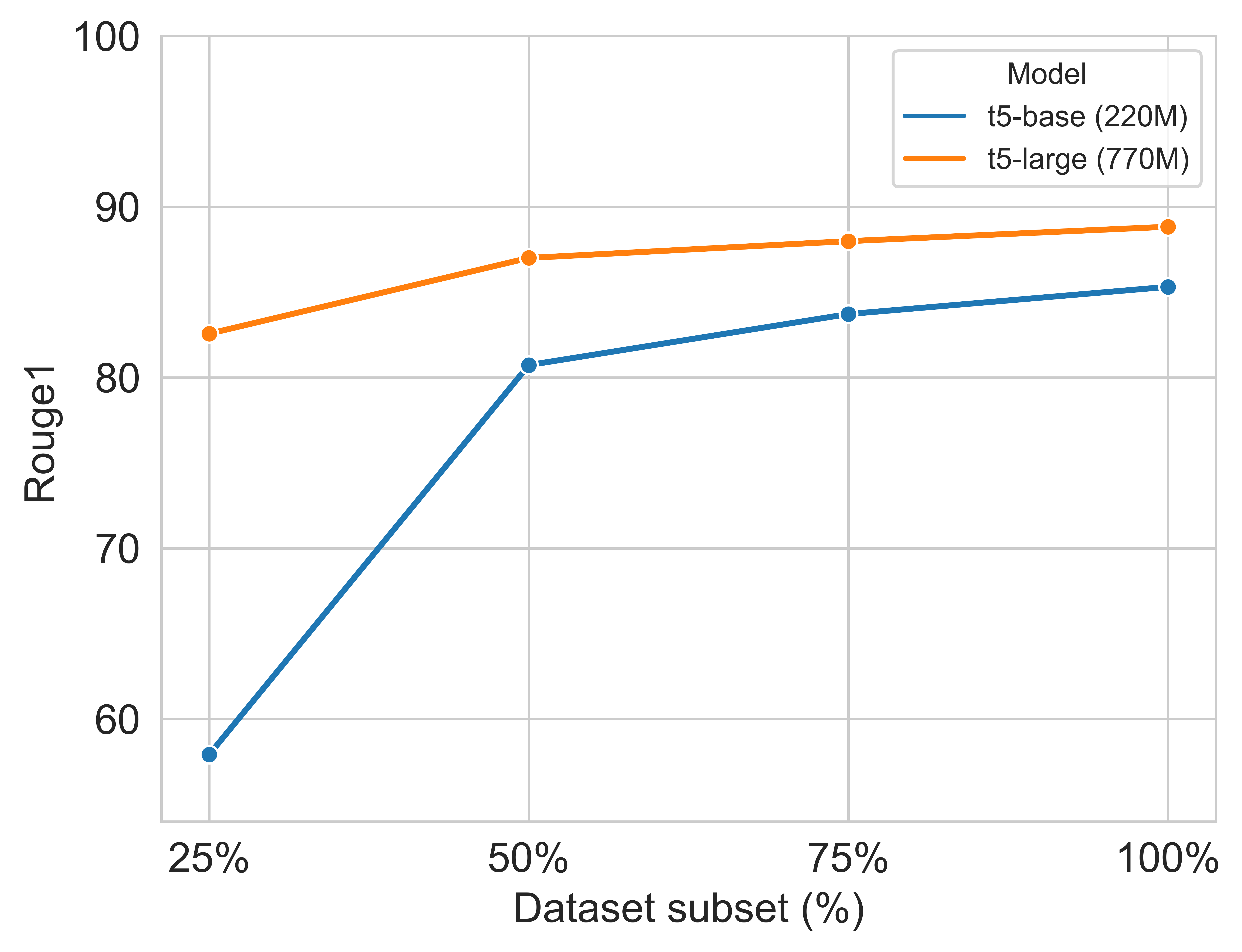}
    \caption{An evaluation of $T5$ models on different subsets of our training data $[25\%, 50\%, 75\%, 100\%]$, as well as different model sizes ($T5_{base}$ and $T5_{large}$). The number of parameters is indicated for each model.}
    \label{fig_learning_curve}
    \vspace{-3mm}
\end{figure}

%% file: figures/figEvaluationInterface.tex
\begin{figure*}[t]
    \centering
    \subcaptionbox{Coverage\label{fig_evaluation_coverage}}{
        \includegraphics[width=0.45\textwidth]{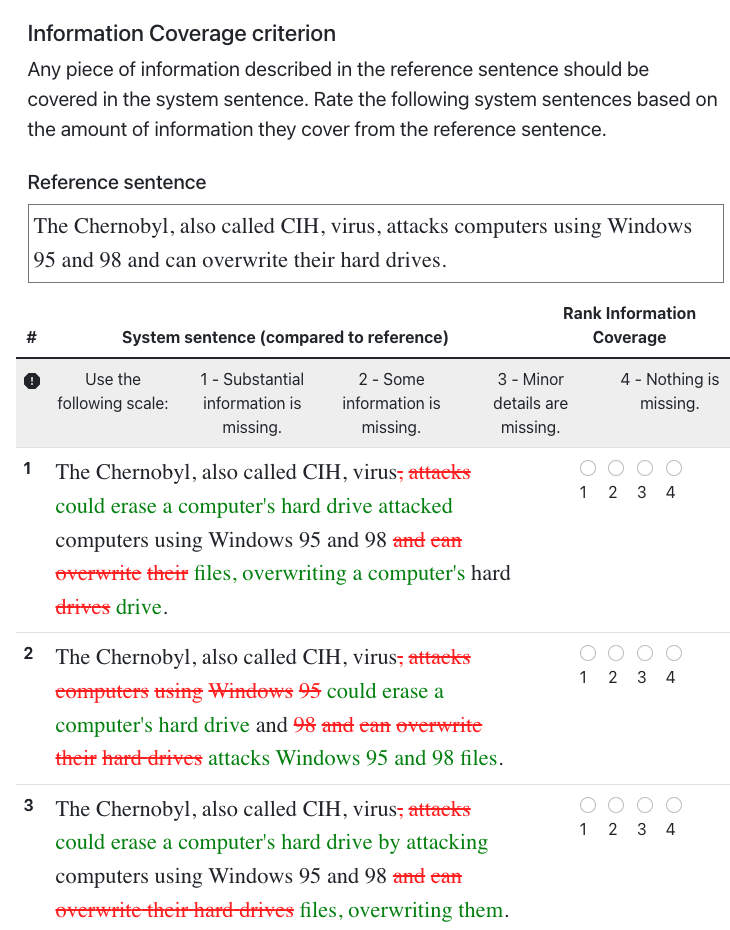}
    }%
    \subcaptionbox{Faithfulness\label{fig_evaluation_faithfulness}}{
        \includegraphics[width=0.45\textwidth]{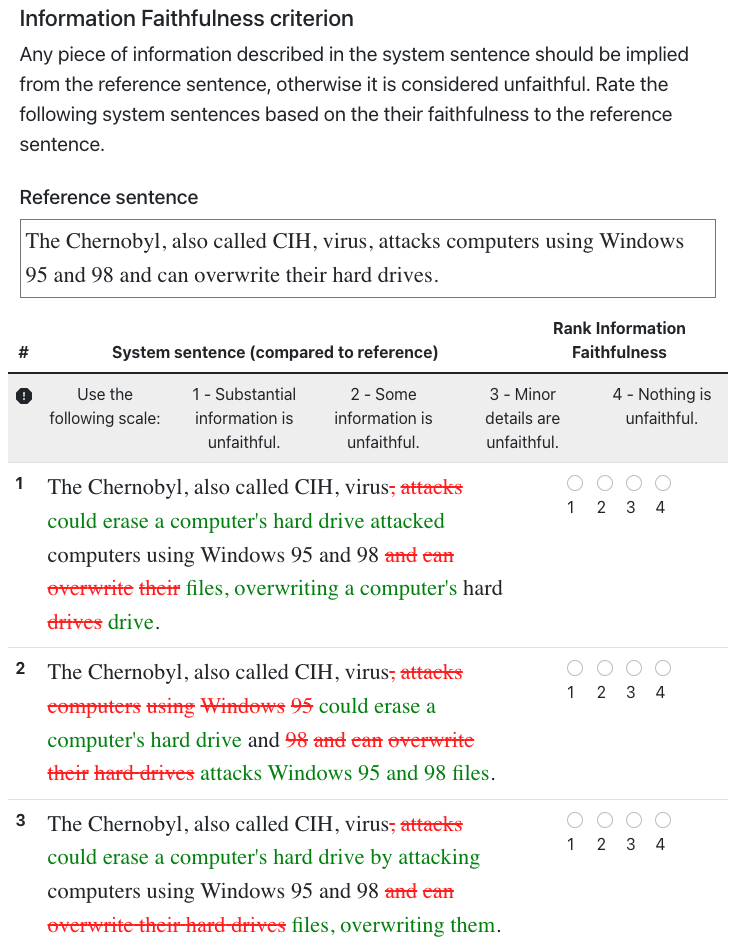}
    }\par\bigskip%
    \subcaptionbox{Repetition\label{fig_evaluation_repetition}}{
    \centering
        \includegraphics[width=0.45\textwidth]{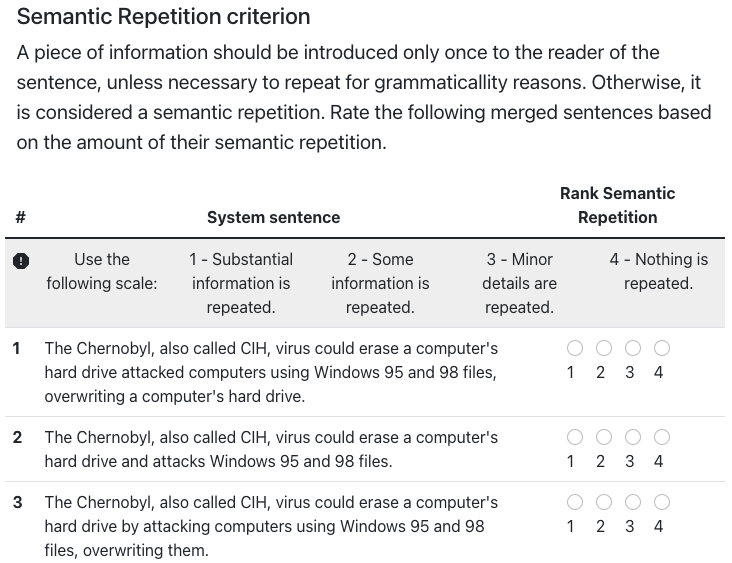}
    }%
    \subcaptionbox{Fluency\label{fig_evaluation_fluency}}{
        \centering
        \includegraphics[width=0.45\textwidth]{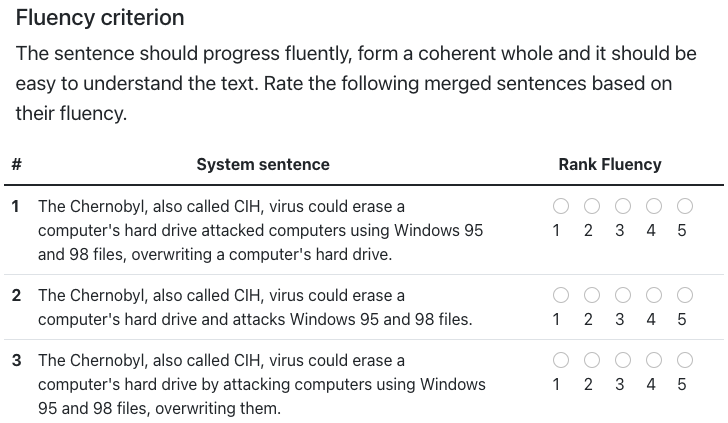}
    }%
    
    \caption{
        The interface used for the evaluation of a predicted sentence union's quality.
    }
    \label{fig_eval_interface}
\end{figure*}

%% file: tables/tabGeneratedExamples.tex
\begin{table*}[t!]
\centering
\resizebox{0.8\textwidth}{!}{%
\begin{tabular}{p{1.8cm}|p{12cm}|p{2.7cm}}
\toprule
      \makecell{Sentence 1} &                                                   French museum officials traveled to New York last month and confirmed the find is indeed the missing Picasso work, which the Centre Georges Pompidou realized was missing from its storerooms in 2001 following a loan request; it was then valued at more than \$2.5 million. &                                                                                                \\[1ex]
      \makecell{Sentence 2} &                                                                                                                                                                       The canvas had been smuggled out of a storeroom of the Centre Georges Pompidou, the Paris museum and arts center, and its whereabouts had not been known. &                                                                                                \\[1ex]
      \makecell{Gold union} & French museum officials traveled to New York last month and confirmed the find is indeed the missing Picasso canvas smuggled out of a storeroom of the Centre Georges Pompidou, the Paris museum and arts center, which realized it was missing in 2001 following a loan request; it was then valued at more than \$2.5 million. &                                                                                                \\[1ex]
      \hline 
\makecell{$T5_{large}$} &           French museum officials traveled to New York last month and confirmed the find is indeed the missing Picasso work, which the Centre Georges Pompidou realized was missing from its storerooms in 2001 following a loan request; it was then valued at more than \$2.5 million, and its whereabouts had not been known. & \makecell{Coverage: 2.0\\[1ex]Faithfulness: 4.0\\[1ex]Repetition: 3.0\\[1ex]Fluency: 3.0\\[1ex]Consolidation: 3.0} \\[1ex]
\makecell{\textsc{Primera}} &                                                   French museum officials traveled to New York last month and confirmed the find is indeed the missing Picasso work, which the Centre Georges Pompidou realized was missing from its storerooms in 2001 following a loan request; it was then valued at more than \$2.5 million. & \makecell{Coverage: 2.0\\[1ex]Faithfulness: 4.0\\[1ex]Repetition: 4.0\\[1ex]Fluency: 5.0\\[1ex]Consolidation: 3.3} \\[1ex]
            \makecell{GPT3} &     French museum officials traveled to New York last month and confirmed that the canvas, which had been smuggled out of a storeroom of the Centre Georges Pompidou in Paris and its whereabouts had not been known since 2001 following a loan request, is indeed the missing Picasso work, valued at more than \$2.5 million. & \makecell{Coverage: 3.0\\[1ex]Faithfulness: 4.0\\[1ex]Repetition: 4.0\\[1ex]Fluency: 2.0\\[1ex]Consolidation: 3.7} \\[1ex]
\bottomrule
\end{tabular}
}
\caption{Examples of predicted union sentences from each baseline system and their corresponding human evaluation.}
\label{tab_predicted_unions}

\end{table*}

%% file: tables/tabErrorAnalysisResults.tex
\begin{table*}[t!]
\centering
\resizebox{\textwidth}{!}{%
\begin{tabular}{l|llll}
\toprule
{} & Coverage & Faithfulness & Repetition &                                                                                                                                                                 \makecell{Subcategory Explanation} \\
Subcategorization                         &          &              &            &                                                                                                                                                                                                    \\
\midrule
Uni-directional entailment                &     17 &          2 &        5 &  \makecell{This includes cases where either the entailing part\\is missing and the entailed part is present in the sentence or both the entailing\\and entailed parts are present in the sentence.} \\
Wrong attachment                          &     13 &         13 &        1 &                                                                                                    \makecell{This includes cases where an argument is attributed to the wrong predicate or entity.} \\
\makecell[l]{Lexical similar \\  but different information} &      8 &            0 &          0 &                                \makecell{This includes cases where information is omitted, and the omitted information\\had a phrase that was lexically similar to a phrase in the other sentence.} \\
Ignores prefix                            &      4 &            0 &          0 &                                                                                            \makecell{This includes cases where the prefix to the sentence in the source is omitted from the union.} \\
Related new information                   &      2 &            0 &          0 &                                                 \makecell{This includes cases where the source sentences contain related\\but different information, and one of them is not included in the union.} \\
Paraphrase                                &      1 &          1 &        5 &                                                                                             \makecell{This includes cases where paraphrased information\\from the source is repeated in the union.} \\
External hallucination                    &        0 &          3 &          0 &                                                                          \makecell{This includes cases where there is information\\in the union that does not originate from the source sentences.} \\
\bottomrule
\end{tabular}
}
\caption{Error analysis based on a sample of 20 erroneous examples, each example analyzed for the 3 system outputs. For each metric, we report the frequency of a subcategory that we suspect is the cause for the error. One representative example from each subcategory is included in Table \ref{tab_error_analysis_examples}.}
\label{tab_error_analysis}

\end{table*}

%% file: tables/tabErrorAnalysisExamples.tex
\begin{table*}[t!]
\centering
\resizebox{\textwidth}{!}{%
\begin{tabular}{p{3cm}|p{12cm}p{6cm}}
\toprule
{} &                                                                                                                                                                                                                                                                                                                                                                                     Prediction &                                                                                                                                                                                                                Explanation \\
Subcategorization                         &                                                                                                                                                                                                                                                                                                                                                                                                &                                                                                                                                                                                                                            \\
\hline
\midrule
External hallucination                    &                                                                                                                                                                                                          Peter Capaldi was revealed as the 12th Doctor of the Doctor Who series during a special live broadcast, with the announcement being made that he had been cast as the 12th Time Lord. &                                                                         The mention of a live broadcast is not part of the source sentences. Interestingly, this is true, which indicates that the model knows this story. \\
Lexical similar but different information &                                                                                                                                                                                                    Sgt. Tim Shields and Attorney-General Wally Oppal announced Wednesday that the RCMP arrested two Bountiful residents, Winston K. Blackmore, 52, and James Oler, 44, on charges of polygamy. &                                                      Source sentence mentioned ``and leaders of a polygamist group''. This was possibly skipped due to the model incorrectly recognizing "polygamy" later as a paraphrase. \\
Uni-directional entailment                &                                                                                                                                                      A strong 6.1-magnitude earthquake which hit the Indonesian northwestern province of Aceh on Tuesday killed a child, injured dozens and destroyed buildings, sparking panic in a region devastated by the quake-triggered tsunami of 2004. &                                                                                  Sentence 2 mentions ``injuring at least 50 people'' which entails ``dozens injured'' in sentence 1, but it is not mentioned in the union. \\
Ignores prefix                            &                                                                                       The 55-year-old Scottish actor Peter Capaldi is officially set to replace exiting star Matt Smith, who announced in June that he was leaving the sci-fi show later this year, as the TARDIS leader, as producer Steven Moffat announced on the live BBC special Doctor Who Live: The Next Doctor Sunday. &                             Ignores the information about it being the 12th doctor, which was mentioned in a sentence prefix: ``Doctor Who has finally selected its 12th doctor: Peter Capaldi is officially set to ...''. \\
Related new information                   &   Industry analysts contacted by eWEEK generally say they believe that Hewlett-Packard's \$13.9 billion acquisition of Electronic Data Systems, which was officially announced on May 13 and is currently being negotiated, is a good move for both companies, although there will be the usual integration snafus such as vendor neutrality issues, business lines, culture shock and layoffs. &    "good move for both companies" and "a deal that could help the world's largest personal computer maker snap up more data management and consulting contracts" are different, and both should be mentioned in the union. \\
Paraphrase                                &                                                                                                                                                                                                                                                                                In France, art dealers are obliged by law to register all purchased art, except those bought at public auction. &  Sentence 1 mentions ``art dealers ... purchases'', and sentence 2 mentions ``dealers ... purchased art''. Since these are paraphrases, the union which repeates both ``art dealers'' and ``purchased art'' is repetitive. \\
Wrong attachment                          &                                                                                                                                     The flight recorder was recovered on November 9 and revealed that the autopilot was disconnected, the descent appeared "controlled," the cockpit turned off both engines, and the elevators were out of unison, something experienced pilots would not do. &                                              ``something experienced pilots would not do'' refers to turning out both engines, not elevators out of unison. This is usually caused by an incorrect merge of the sentences. \\
\bottomrule
\end{tabular}
}
\caption{Examples for the subcategories we devised during the model error analysis, which we suspect are are the cause for the error.}
\label{tab_error_analysis_examples}

\end{table*}